\documentclass[runningheads]{llncs}

\usepackage[T1]{fontenc}
\usepackage{graphicx}
\usepackage{color}

\usepackage{url}
\usepackage{cite}
\usepackage{soul}
\usepackage{csquotes}
\usepackage{hyperref}
\usepackage{amssymb, amsmath}
\usepackage{marvosym} 

\begin{document}
    \title{Beyond Predictive Fairness: Quantifying Attribution Consistency Across Demographic Groups in Diabetic Retinopathy Screening}
    
    \titlerunning{Quantifying Attribution Consistency Across Demographic Groups}
    
     \author{
        Kerol Djoumessi\inst{1}\textsuperscript{(\Letter)}\orcidID{0009-0004-1548-9758} \and Philipp Berens\inst{1,2}\orcidID{0000-0002-0199-4727}
    }
     
     \authorrunning{K. Djoumessi \& P. Berens}
    
    \institute{
        Hertie Institute for AI in Brain Health, University of T\unexpanded{\"u}bingen, Germany
        \email{kerol.djoumessi-donteu@uni-tuebingen.de}\\ \url{https://hertie.ai/} \and
        T\unexpanded{\"u}bingen AI Center, University of T\unexpanded{\"u}bingen, Germany\\
    }
    \maketitle             
    
    \begin{abstract}
        Fairness in medical imaging is commonly evaluated through subgroup performance metrics, yet it remains unclear whether models rely on consistent visual evidence across demographic groups. This work introduces the Explanation Consistency Score (ECS), a fairness-aware metric based on Jensen–Shannon divergence that quantifies the similarity of attribution maps across subgroups. Using diabetic retinopathy screening as a case study, ECS is evaluated globally and within disease severity. Experiments reveal that while predictive performance differs across ethnic groups, explanation consistency remains relatively high and shows no significant association with performance disparities. These findings suggest that predictive fairness and explanation consistency capture complementary dimensions of model behavior, motivating fairness evaluations that extend beyond predictive performance.
        
    \end{abstract}
    
    \section{Introduction}
        Deep learning systems have achieved strong performance in automated medical image diagnosis across a wide range of clinical tasks \cite{vrudhula2024machine}, including diabetic retinopathy (DR) screening from retinal fundus images \cite{alyoubi2020diabetic, djoumessi2025inherently}. Alongside these advances, fairness and explainability have become increasingly important concerns in medical imaging due to the reported performance disparities across demographic subgroups \cite{alloula2024biases, vrudhula2024machine}. Existing fairness studies primarily evaluate subgroup disparities using predictive metrics such as sensitivity, specificity, and area under the receiver operating characteristic curve (AUC) \cite{larrazabal2020gender, mehta2024evaluating}. While such evaluations quantify differences in predictive outcomes, they provide limited insight into whether models rely on similar visual evidence across patient groups.

        Recent studies show that medical imaging models can encode demographic information despite being trained for clinical tasks \cite{yang2024limits, muller2022generative}---in ophthalmology, demographic attributes can even be inferred directly from fundus images \cite{muller2022generative, ilanchezian2021interpretable}. Consequently, two models may show similar subgroup performance while relying on different visual evidence, or vice versa, making predictive metrics alone insufficient for trustworthy deployment.

        Explainability methods such as Class Activation Mapping (CAM) are widely used in ophthalmology to improve interpretability by highlighting retinal regions contributing to model predictions \cite{ayhan2022clinical, omeiza2019smooth, wang2020score}. These methods have been used to verify whether models attend to clinically relevant retinal structures and lesions, thereby supporting the interpretation of automated screening systems.
        Despite known limitations and sensitivity to model architecture \cite{adebayo2018sanity, wang2020score}, attribution maps remain one of the most widely adopted approaches for understanding model behavior in medical imaging \cite{bhati2024survey, ayhan2022clinical}. However, existing studies have primarily leveraged explainability qualitatively to inspect subgroup failures, identify potential sources of bias, or visualize fairness-related concerns \cite{zhao2023fairness, zhou2020towards}. Quantitative analyses of attribution differences across demographic groups remain limited \cite{pfohl2026understanding}, and their relationship with predictive fairness is not yet well understood. This raises an important question: \emph{do predictive disparities necessarily imply differences in model reasoning across demographic groups?}

        In this work, we investigate the demographic consistency of visual attribution patterns in early diabetic retinopathy detection using retinal fundus images from the EyePACS dataset. 
        An \emph{Explanation Consistency Score (ECS)} is introduced as a Jensen--Shannon divergence-based metric to quantify attribution similarity across demographic groups. 
        To account for potential confounding effects arising from differences in disease severity, a conditional formulation of ECS is additionally proposed. By jointly analyzing predictive performance and explanation consistency across demographic groups, this work provides a quantitative framework for fairness-aware attribution analysis in medical imaging.

    \section{Methods}
        \subsection{Dataset and Demographic Subgroups}
            Experiments were conducted on the EyePACS diabetic retinopathy dataset \cite{cuadros2009eyepacs}, which provides retinal fundus images, DR grades (0--4), and patient ethnicity metadata. Following the provided quality-control protocol, 90,595 good-quality images were retained (Tab.~\ref{tab:dataset}). 
            Data were split at the patient level into $70\%$ training, $15\%$ validation, and $15\%$ test sets using iterative multi-label stratification based on ethnicity and DR grade to preserve subgroup and disease distributions.

            The fairness analysis focused on the five largest ethnicity groups: Latin American, African Descent, Indian Origin, Caucasian, and Asian. Samples with unspecified ethnicity or underrepresented groups were retained for training but excluded from subgroup analyses due to missing or limited data.
            
            \begin{table}[h] 
                \caption{Demographic distribution and DR grade composition of the EyePACS cohort.}
                \centering
                \begin{tabular}{l|c|c|c|c|c|c|c|c}
                    \hline
                   \text{Ethnicity }  & \text{\# Samples} & \text{ DR0 } & \text{ DR1 } & \text{ DR2 } & \text{ DR3 } & \text{ DR4 } &\text{ DR1-4 } &\text{ Prev.} \\
                   \hline
                   Latin American & $53,960$ & $42,155$ & $5,787$ & $5,502$ & $285$ & $231$ & $11,805$ & $21.9\%$ \\   
                   African descent & $5,097$ & $3,597$ & $747$ & $663$ & $40$ & $50$ & $1,500$ & $29.4\%$ \\  
                   Indian origin & $3,341$ & $1,822$ & $472$ & $930$ & $51$ & $66$ & $1,519$ & $45.5\%$ \\ 
                   Caucasian & $9,542$ & $7,818$ & $794$ & $884$ & $24$ & $22$ & $1,724$ & $18.1\%$\\ 
                   Asian & $3,995$ & $3,335$ & $321$ & $314$ & $4$ & $21$ & $660$ & $16.5\%$\\ 
                   \hline
                   Not specified & $12,253$ & $9,825$ & $1,228$ & $1,114$ & $60$ & $26$ & $2,428$ & $19.8\%$\\ 
                   Underrepresented & $2,407$ & $1,816$ & $244$ & $324$ & $17$ & $6$ & $591$ & $24.5\%$ \\ 
                   \hline                    
                \end{tabular}
                \label{tab:dataset}
            \end{table}

        \subsection{DR classification model}
            A ResNet-50 \cite{he2016deep} was trained for early diabetic retinopathy (DR) detection by distinguishing healthy eyes (grade 0) from eyes exhibiting any stage of DR (grade 1--4). This binary setting was selected because early disease detection requires identifying subtle retinal lesions \cite{djoumessi2025inherently} and therefore provides a suitable benchmark for studying potential demographic differences in model explanations.

        \subsection{Attribution Map Generation}
            \label{sec:attrib}
            To analyze the visual evidence underlying model predictions, attribution maps were generated using SmoothGradCAM++ \cite{omeiza2019smooth} and Score-CAM \cite{wang2020score}, representing gradient-based and gradient-free attribution methods, respectively. Using both methods enables evaluating the robustness of the proposed explanation consistency analysis across different attribution mechanisms.
            Each map \(A \in \mathbb{R}^{N \times M}\) was normalized to sum to one (\( \tilde{A}_{i,j} = A_{i,j} / \sum_{u,v} A_{u,v} \)), yielding a spatial probability distribution.

        \subsection{Explanation Consistency Score}
            To quantify attribution consistency across demographic groups, an \emph{Explanation Consistency Score} (ECS) is introduced. For a demographic group \(g\), the mean attribution distribution is defined as 
            \[ \mu_g = \mathbb{E}_{x \sim P(x|g)}  [\tilde{A}(x)] \approx \frac{1}{|D_g|} \sum_{x \in D_g} \tilde{A}(x), \]

            where \(D_g\) denotes the set of samples belonging to group \(g\). 
            Attribution distributions are compared using Jensen--Shannon (JS) divergence \cite{menendez1997jensen}, a symmetric and bounded measure of similarity between probability distributions. The pairwise ECS between demographic groups \(g_1\) and \(g_2\) is defined as
            \[ \mathrm{ECS}(g_1,g_2) = 1-\mathrm{JS} (\mu_{g_1}\|\mu_{g_2}). \]

            ECS ranges from 0 to 1, where larger values indicate greater similarity between attribution distributions. 
            To summarize the consistency of a demographic group relative to all remaining groups, a multi-group ECS is computed as
            \[ \mathrm{AvgECS}(g) = \frac{1}{| \mathcal{G}|-1} \sum_{g' \neq g} \mathrm{ECS}(g,g'), \]

            where \(\mathcal{G}\) denotes the set of demographic groups.
            Disease severity may act as a confounding factor because attribution patterns can vary across DR grades. To control this effect, ECS is additionally computed conditionally on DR grade. For demographic group \(g\) and DR grade \(y\), the conditional attribution distribution is defined as
            \[ \mu_g^{(y)} = \mathbb{E}_{x \sim P(x|g,y)} [\tilde{A}(x)]. \]

            The conditional ECS for grade $y$ is then
            \[ \mathrm{ECS}^{(y)}(g_1,g_2) = 1- \mathrm{JS} \left( \mu_{g_1}^{(y)} \| \mu_{g_2}^{(y)} \right), \]

            and the final conditional score is obtained by averaging across grades:
            \[ \mathrm{CondECS}(g_1,g_2) = \frac{1}{|\mathcal{Y}|} \sum_{y \in \mathcal{Y}} \mathrm{ECS}^{(y)}(g_1,g_2). \]

            In practice, conditional analyses were restricted to DR grades 1 and 2, corresponding to the early disease stages considered in this study.
        
    \section{Experiments and Results}
        \subsection{Experimental Setup}
            \paragraph{\textbf{Data Preprocessing and Augmentation.}} 
            Images were preprocessed by circular field-of-view cropping \cite{Mueller_fundus_circle_cropping}, resized to $512 \times 512$, and normalised using training set statistics. Augmentation followed \cite{huang2023identifying},  including random cropping, color jittering, and rotation.
            
            \paragraph{\textbf{Model Training.}} Following prior work \cite{huang2023identifying, djoumessi2025inherently}, the model\footnote{The code is available at: \url{https://github.com/kdjoumessi/Fairness-ECS}} was initialized with ImageNet-pretrained weights and optimized using cross-entropy loss. 
            Training used SGD with an initial learning rate of \(10^{-3}\) and a clipped cosine schedule, selecting the model with highest validation AUC after 100 epochs.            
            \paragraph{\textbf{Attribution Processing.}} 
            Attribution maps were generated using the TorchCAM library \cite{torcham2020} at the resulting native low resolution ($16 \times 16$ pixels), normalized for all consistency analyses, and upsampled for qualitative visualization.

        \subsection{Predictive Performance Across Demographic groups}
            Subgroup predictive performance was assessed using AUC, sensitivity, and specificity (Tab.\,\ref{tab:predictive_performance}). The latter two directly reflecting clinical screening relevance.
            \begin{table}[t] 
                \centering
                \caption{Predictive performance and attribution consistency across demographic groups. Results are reported as mean $\pm$ standard deviation over three random seeds.}
                \footnotesize
                \setlength \tabcolsep {1pt} 
                \begin{tabular}{l|c|c|c|c|c|c|c}
                    \hline
                    & \multicolumn{3}{c|}{Predictive performance} & \multicolumn{4}{c}{Explanation consistency} \\
                    \hline
                   Ethnicity & \textbf{ AUC } & \textbf{ Sens. } & \textbf{ Spec. } & \textbf{ECS-SC} & \textbf{WG-ECS} & \textbf{ECS-SG} & \textbf{DR1-ECS}\\
                    \hline
                    Latin American  & $.85 \pm .002$ & $.57 \pm .013$ & $.97 \pm .004$ & $.92 \pm .002$ & $.53 \pm .08$ & $.88 \pm .010$  & $.87 \pm 006$ \\
                    African descent & $.88 \pm .002$ & $.64 \pm .023$ & $.97 \pm .002$ & $.91 \pm 005$ & $.56 \pm .08$ & $.88 \pm .008$ & $.85 \pm .014$ \\
                    Indian origin & $.92 \pm .001$ & $.77 \pm .022$ & $.94 \pm .007$ & $.90 \pm 006$ & $.57 \pm .08$ & $.86 \pm .010$ & $.85 \pm 009$ \\
                    Caucasian & $.83 \pm .003$ & $.54 \pm .018$ & $.97 \pm .004$ & $.91 \pm 004$ & $.57 \pm .07$ & $.86 \pm .014$ & $.85 \pm 008$ \\
                    Asian  & $.90 \pm .006$ & $.68 \pm .023$ & $.97 \pm .005$ & $.89 \pm .004$ & $.54 \pm .08$ & $.85 \pm .008$ & $.79 \pm .011$ \\
                    \hline
                    All  & $.86 \pm .001$ & $.59 \pm 015$ & $.97 \pm 004$ & $-$ & $-$ & $-$ & $-$ \\ 
                    \hline
                \end{tabular}
                \label{tab:predictive_performance}
            \end{table}
            
            Performance differences were observed across demographic groups. Patients of Indian origin achieved the highest AUC ($0.92 \pm 0.001$) and sensitivity ($0.83 \pm 0.003$), whereas Caucasian patients exhibited the lowest AUC ($0.83 \pm 0.003$) and sensitivity ($0.54 \pm 0.018$), corresponding to a sensitivity gap of $23\%$. In contrast, specificity remained relatively stable across groups ($0.94$--$0.97$), indicating that subgroup differences primarily arise from the detection of positive DR cases rather than healthy patients. 

            Interestingly, predictive performance did not follow subgroup size (Tab.\,\ref{tab:dataset}). Although Latin American patients represent the largest demographic group, they did not achieve the strongest performance. Conversely, Indian-origin and Asian patients, among the smallest groups considered, obtained the highest AUC and sensitivity values. 
            Neither subgroup size nor disease prevalence alone explains the observed differences: the Asian subgroup achieved strong performance despite the lowest prevalence, and Indian-origin patients outperformed the much larger Latin American group.
            Low standard deviations across all metrics indicate that these trends were consistent across random initializations.
            These findings reveal measurable predictive disparities across ethnicities and motivate the subsequent analysis of whether such differences are reflected in model attribution behavior.
            
        \subsection{Attribution Consistency Across Demographic Groups}        
\begin{figure}[t]
                \centering
                \includegraphics[width=\textwidth]{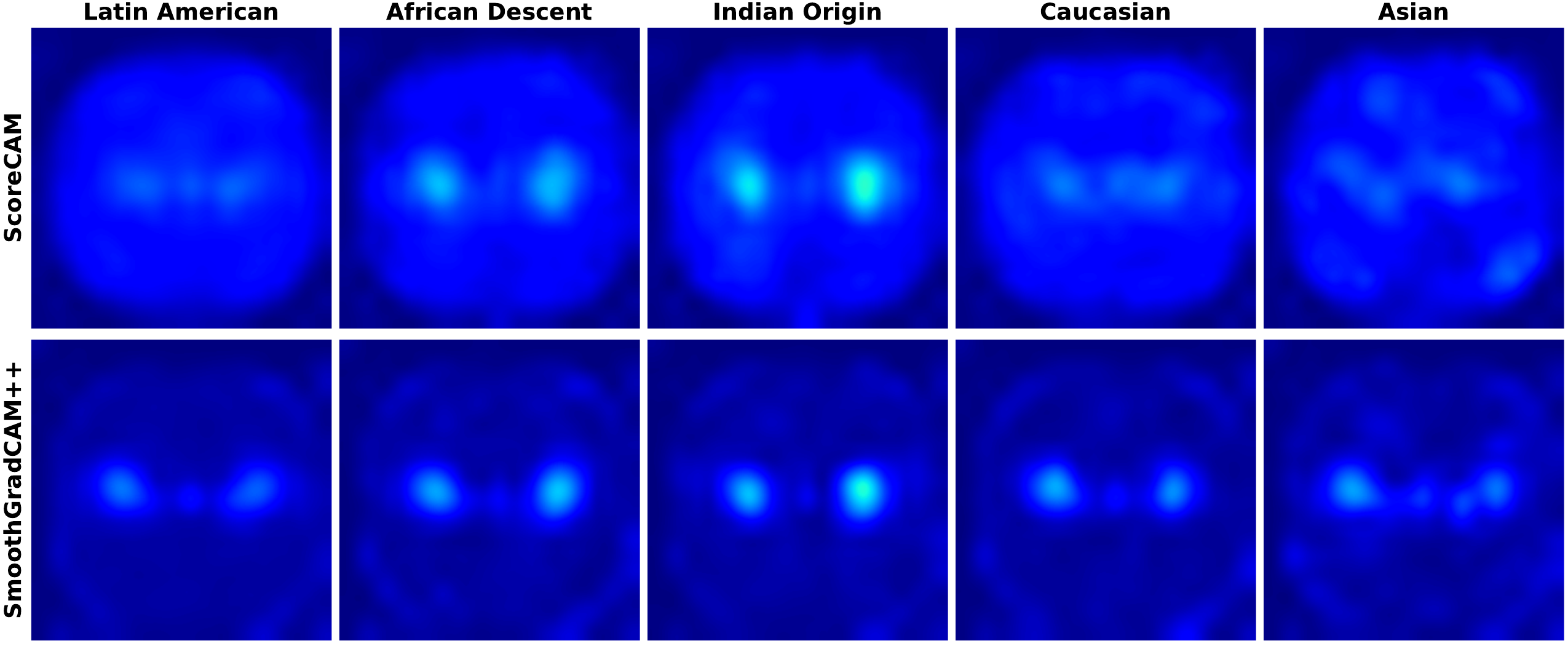}
                \caption{\textbf{Explanation consistency across demographic groups.} Average ScoreCAM and SmoothGradCAM++ attribution maps for correctly classified DR-positive samples across demographic groups. Similar attention patterns are observed across ethnicities, indicating high explanation consistency.}
                \label{fig1:ecs}
            \end{figure}

            To assess whether model explanations differ across demographic groups, the proposed Explanation Consistency Score (ECS) for a given model was computed from correctly classified test samples using ScoreCAM (ECS-SC) and smoothGradCAM++ (ECS-SG). This design choice---restricting to correct predictions---isolates explanation consistency under successful decision-making, as misclassified examples may exhibit heterogeneous failure modes that complicate cross-group comparisons. Multi-group ECS values are reported in Table \ref{tab:predictive_performance}, while average attribution maps are shown in Figure \ref{fig1:ecs}.             
            High explanation consistency was observed across all demographic groups. ECS values ranged from $0.89$ to $0.92$ for ScoreCAM and from $0.85$ to $0.88$ for SmoothGradCAM++, indicating that attribution patterns remained largely similar across ethnicities. These findings are consistent with the qualitative visualizations (Fig.\,\ref{fig1:ecs}), where both attribution methods consistently focus on similar retinal regions despite minor differences in attribution concentration and spatial distribution.

            Compared with predictive performance, attribution consistency exhibited substantially lower variability across groups. While sensitivity varies by $23\%$, ECS varies by only $0.03$ for both attribution methods. Notably, the subgroup achieving the highest predictive performance (Indian origin) did not exhibit the highest ECS, whereas groups with lower predictive performance, such as Caucasian patients, remained highly consistent with the attribution patterns observed in other groups. These findings suggest that predictive disparities are not necessarily accompanied by comparable differences in attribution behavior.            
            To further investigate this relationship, Spearman correlations ($\rho$) were computed between subgroup performance and ECS. 
            Moderate negative correlations were observed for both AUC and sensitivity ($\rho= - 0.80$ for ScoreCAM, $\rho= - 0.60$ for SmoothGradCAM++), though neither reached statistical significance---unsurprising given the limited number of demographic groups ($n=5$), preventing meaningful evidence.

            To assess whether group-level ECS values could be explained by differences in within-group attribution variability, the similarity between each attribution map and its corresponding group-average attribution map was additionally measured. Within-group consistency (WG.ECS) was comparable across demographic groups (ScoreCAM: $0.53–0.57$; SmoothGradCAM++: $0.43–0.50$), indicating comparable levels of attribution variability and supporting the validity of group-level ECS comparisons. This 
            suggests that the high ECS values are unlikely to be driven by subgroup-specific differences in attribution variability. 
            
        \subsection{Conditional Attribution Consistency Analysis}
            \begin{figure}[t]
                \centering
                \includegraphics[width=\textwidth]{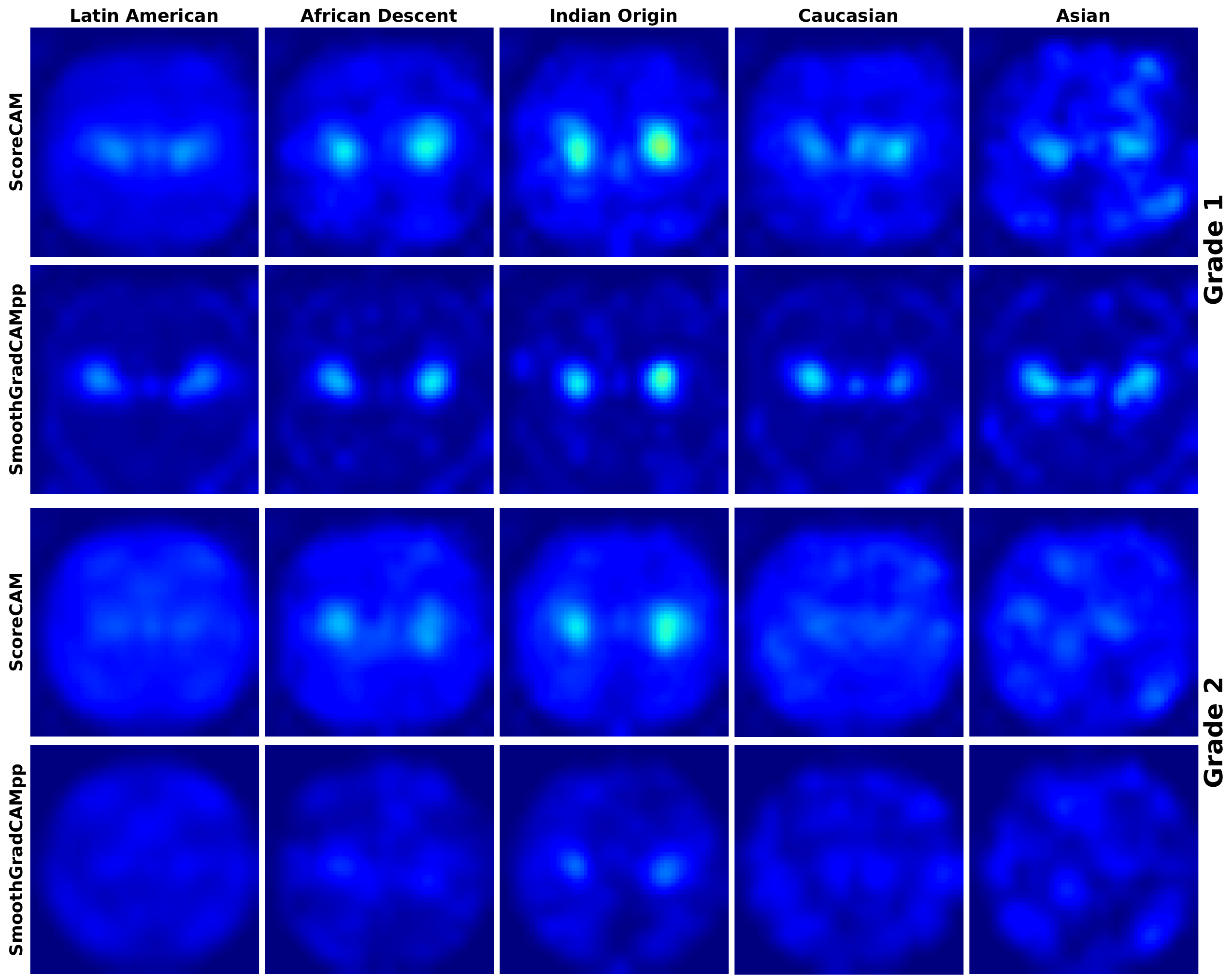}
                \caption{\textbf{Conditional attribution consistency across demographic groups.}
                Average attribution maps for DR grades 1 (top two rows) and 2 (bottom two rows) across demographic groups using ScoreCAM and SmoothGradCAM++. Attribution patterns remain broadly consistent within each disease grade, supporting the robustness of the proposed ECS analysis after controlling for disease severity. }
                \label{fig2:conditional_ecs}
                \vspace{-.25cm}
            \end{figure}
            
            Differences in disease prevalence and severity across demographic groups may influence attribution patterns and potentially confound the interpretation of explanation consistency. To control for this effect, conditional ECS was computed separately for correctly classified DR grades 1 and 2, thereby comparing attribution maps only among patients with the same disease severity. 
            
            High explanation consistency was preserved after conditioning on disease severity (Fig.~\ref{fig2:conditional_ecs}). For ScoreCAM, multi-group ECS ranged from $0.79$ to $0.87$ for grade 1 (DR1-ECS, Tab.\,\ref{tab:predictive_performance}) and from $0.87$ to $0.90$ for grade 2 across demographic groups. Similarly, trends were observed for SmoothGradCAM++ produced ECS values ranging between $0.77$--$0.82$ for grade 1 and $0.81$--$0.86$ for grade 2. The highest consistency was generally observed for Latin American patients, whereas Asian patients exhibited the lowest ECS values across both grades and attribution methods.
            Across all demographic groups and attribution methods, ECS was consistently higher for grade 2 than for grade 1. For example, the ScoreCAM-based ECS of Asian patients increased from $0.79 \pm 0,01$ (grade 1) to $0.87 \pm 0.002$ (grade 2). This trend suggests that attribution patterns become more stable as diabetic retinopathy lesions become more visually apparent, whereas early-stage disease (grade 1) is associated with greater attribution variability. 

            Importantly, high ECS values persisted after controlling for disease severity, indicating that the consistency observed in the global analysis (Fig.~\ref{fig1:ecs}) is not solely explained by differences in disease prevalence or grade distribution across across demographic groups. Overall, these findings provide further evidence that the model relies on broadly similar visual evidence across demographic groups even when comparisons are restricted to patients with the same disease stage. 

    \section{Discussion and Conclusion}
        This work introduced the Explanation Consistency Score (ECS) to quantify the similarity of attribution patterns across demographic groups in diabetic retinopathy screening and investigated whether demographic differences in predictive performance can be explained by differences in model attribution patterns. Although predictive performance varied across ethnicities, attribution consistency remained uniformly high for both ScoreCAM and SmoothGradCAM++, indicating that predictive fairness and explanation consistency capture complementary rather than equivalent aspects of model behavior. In particular, performance disparities were substantially larger than attribution disparities, and no significant association was observed between ECS and subgroup predictive performance.

        Several observations may help explain this decoupling. First, subgroup performance did not appear to be determined solely by sample size: the Indian-origin subgroup achieved the highest AUC and sensitivity despite being among the smallest groups, whereas the largest subgroup did not achieve the strongest performance. Second, disease prevalence may partially contribute to these differences, as higher prevalence generally leads to higher sensitivity, although the strong performance of the Asian subgroup despite its low prevalence suggests that prevalence alone is insufficient to explain subgroup variation. These findings point toward additional factors, such as differences in disease manifestation, image characteristics, or subgroup-specific data distributions \cite{muller2022generative, alloula2024biases, mehta2024evaluating}.
        At the same time, attribution consistency remained high across all groups, including those with lower predictive performance, suggesting that the model relies on broadly similar visual evidence despite differences in predictive outcomes. The conditional analysis further supported this conclusion, as high ECS values persisted after controlling for disease severity. Interestingly, grade-2 images exhibited higher ECS than grade-1 images, suggesting that more advanced disease stages produce more stable attribution patterns. 

        Several limitations should be acknowledged. First, ECS was computed on correctly classified samples to characterize attribution consistency under successful predictions. Consequently, subgroup-specific failure modes remain unexplored and should be investigated in future work. 
        Second, ECS is a group-level measure based on mean attribution distributions. Although an additional within-group analysis revealed comparable attribution variability across demographic groups, suggesting that the observed ECS values are not merely averaging artifacts, ECS may not fully capture individual-level explanation variability. Future work could complement ECS with image-level analyses of attribution variability.
        Third, attribution maps were analyzed at their native CAM resolution, which may limit sensitivity to fine-grained spatial differences. Fourth, ECS relies on Jensen--Shannon divergence; alternative similarity measures such as cosine similarity, Wasserstein distance, or structural similarity may capture complementary notions of attribution similarity \cite{levy2025guide}. Finally, the study was conducted on a single dataset and focused exclusively on ethnicity, using posthoc methods, motivating future evaluation across additional modalities, datasets, protected attributes, and self-explainable models \cite{djoumessi2024actually, djoumessi2026softcam, donteu2023sparse}. 
        
    \begin{credits}
        \subsubsection{\ackname} 
            This project was supported by the Hertie Foundation and by the Deutsche Forschungsgemeinschaft under Germany’s Excellence Strategy with the Excellence Cluster 2064 “Machine Learning – New Perspectives for Science”, and a regular grant (project number 571331899).
        
        \subsubsection{\discintname} 
            The authors declare no competing interests.
    \end{credits}
    
     \bibliographystyle{splncs04}
     \bibliography{biblio}
\end{document}